\documentclass[letterpaper, 10 pt, conference]{ieeeconf}  

\IEEEoverridecommandlockouts                              

\usepackage{graphicx}
\usepackage{caption}
\usepackage{caption}
\usepackage{subcaption}
\usepackage{amsmath,amssymb,enumerate}

\usepackage{amsthm}
\usepackage{setspace}
\usepackage{amsmath,amssymb,enumerate}
\usepackage{booktabs}
\usepackage[usenames,dvipsnames,svgnames,table]{xcolor}
\usepackage{mathtools}
\usepackage{algorithm, algorithmicx, algpseudocode}

\usepackage{blindtext}
\usepackage{gensymb}
\usepackage{xparse}
\usepackage{lipsum}
\usepackage{mathrsfs}
\usepackage[mathscr]{euscript}
\usepackage{times}
\usepackage[noadjust]{cite} 
\usepackage{multicol}
\usepackage{multirow}
\definecolor{darkgreen}{rgb}{0,0.6,0}
\usepackage[bookmarks=true,colorlinks=true,pdfpagemode=UseNone,citecolor=black,linkcolor=black,urlcolor=BrickRed]{hyperref}
\usepackage{amsfonts}
\usepackage{cleveref}
\usepackage[utf8]{inputenc}
\usepackage[T1]{fontenc}
\usepackage{textcomp}
\usepackage{arydshln}
\usepackage{tabu}
\usepackage{cuted}
\usepackage{balance}

\DeclareMathOperator*{\argmax}{arg\,max}

\definecolor{note}{rgb}{0.1,0.1,1}
\definecolor{rephase}{rgb}{0.15,0.7,0.15}
\definecolor{bag}{rgb}{0.6,0.6,0.2}

\makeatletter
\renewcommand*\env@matrix[1][c]{\hskip -\arraycolsep
  \let\@ifnextchar\new@ifnextchar
  \array{*\c@MaxMatrixCols #1}}
\makeatother

\newcommand{\m}{\mathop{\mathrm{m}}}

\DeclareDocumentCommand{\vector}{ O{} }{\mathrm{vec}(#1)}

\makeatletter
\newcommand{\mathleft}{\@fleqntrue\@mathmargin0pt}
\newcommand{\mathcenter}{\@fleqnfalse}
\makeatother

\title{\LARGE \bf SoLo T-DIRL: Socially-Aware Dynamic Local Planner based on Trajectory-Ranked Deep Inverse Reinforcement Learning
}

\author{Yifan Xu, Theodor Chakhachiro, Tribhi Kathuria, and Maani Ghaffari%
\thanks{Funding for M. Ghaffari was 
provided by NSF Award No. 2118818.}
\thanks{The authors are with the University of Michigan, Ann Arbor, MI 48109, USA. {\tt\small\{yfx, teochiro, tribhi, maanigj\}@umich.edu}}%
}

\begin{document}

\maketitle
\thispagestyle{empty}
\pagestyle{empty}

\begin{abstract}

This work proposes a new framework for a socially-aware dynamic local planner in crowded environments by building on the recently proposed Trajectory-ranked Maximum Entropy Deep Inverse Reinforcement Learning \mbox{(T-MEDIRL)}. To address the social navigation problem, our multi-modal learning planner explicitly considers social interaction factors, as well as social-awareness factors into T-MEDIRL pipeline to learn a reward function from human demonstrations. Moreover, we propose a novel trajectory ranking score using the sudden velocity change of pedestrians around the robot to address the sub-optimality in human demonstrations.
Our evaluation shows that this method can successfully make a robot navigate in a crowded social environment and outperforms the state-of-art social navigation methods in terms of the success rate, navigation time, and invasion rate.

\end{abstract}

\IEEEpeerreviewmaketitle

\section{Introduction}
\label{sec:intro}

Robotic technology has enabled the development of Socially Assistive Robots (SARs) to assist humans in various social contexts ~\cite{SARs}. 

Recently, service and guide robots have been deployed in museums~\cite{museum_robot}, shopping malls~\cite{shopping_mall} and airports~\cite{airport_robot}, and are becoming an irreplaceable part of our daily life. 
Assistive robots are required to navigate in public spaces among people in a safe and socially acceptable manner. This problem is known in literature as social navigation~\cite{Eval_SARN,social_aware_navigation}.

The main challenge in social navigation is to infer the underlying social dynamics of humans in the scene~\cite{SORO}. This problem is challenging because people's walking speed and direction can change, and it is hard to quantify different social dynamics and integrate them into the robot planning pipeline. In other words, the robot does not have access to the scene context and such information is not directly observable by its onboard sensors.

The work of~\cite{Fox1997TheDW,social_force_planner,SAPP,CADRL} either treat everything on the way as obstacles to be avoided or only take into account social dynamics navigation behaviors but never both.

These single-modal methods do not support navigation in complex, dynamic social environments. 
Inverse Reinforcement Learning (IRL) in socially-aware navigation has been applied to indoor service robots~\cite{SORL,SORO}. Different from classical methods, IRL-based planners can exploit multi-modal reward function learning and use human demonstrations to train the reward network; thereby, overcoming the inconvenience and inefficiency of hand-designed reward functions.

However, the state-of-the-art IRL-based social navigation methods such as \cite{SORL,SORO} assume the demonstrations are optimal. Because people cannot access all information around the robot and perfectly infer pedestrians' intentions, this assumption is not always satisfied in practice. Also, even in some less crowded situations ensuring optimality in the real world is still challenging, and the performance of these methods often does not outperform the demonstrations.

\begin{figure}[t]
    \centering
    \includegraphics[width =1.0 \columnwidth]{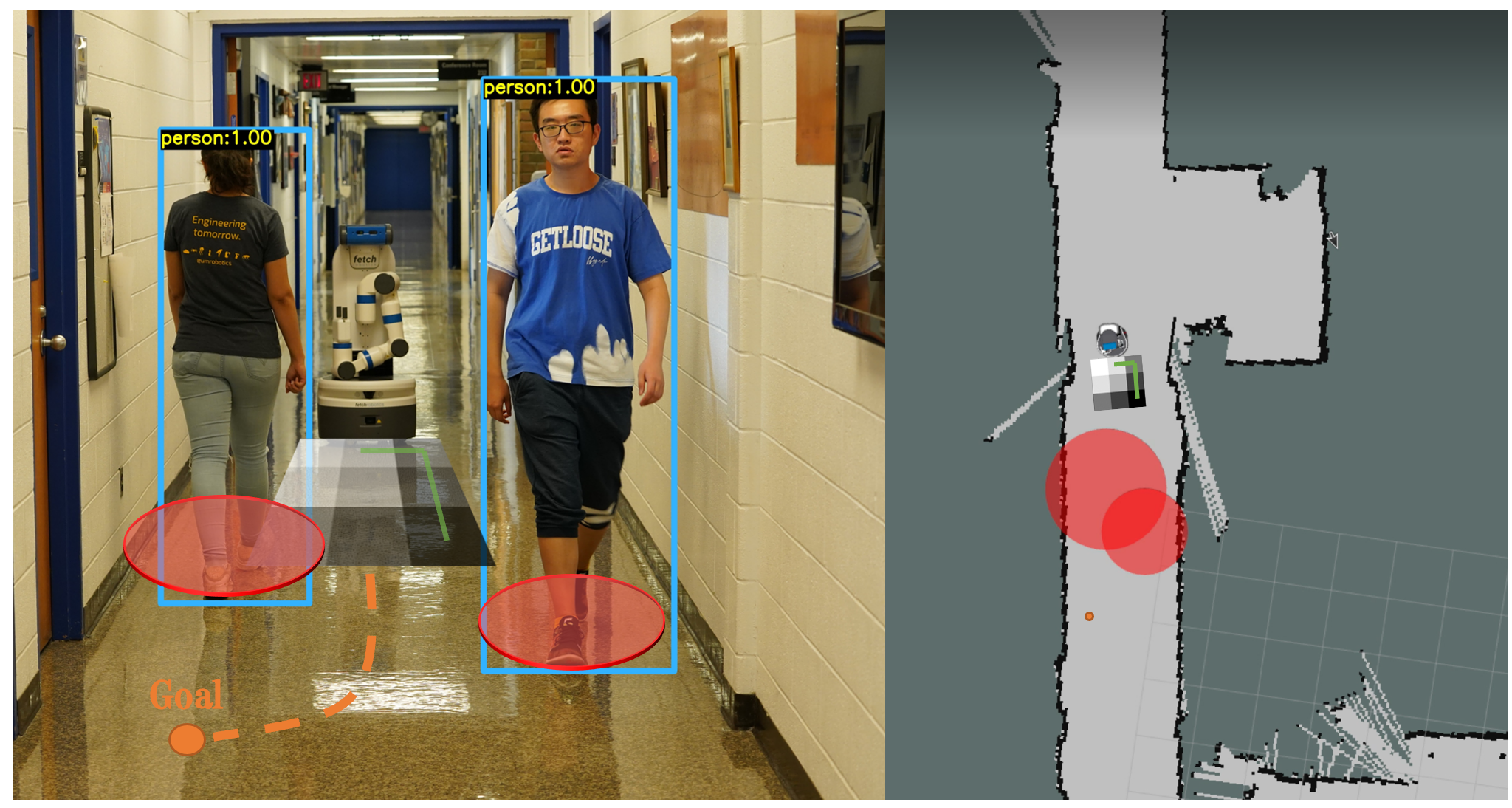}
    \caption{The left figure shows our Fetch robot navigating in the corridor of the Naval Architecture and Marine Engineering building at the University of Michigan while the right figure is the corresponding RViz visualization of the augmented map. Red circles represent the personal space of the two people. Blue bounding boxes are the output detection from YOLOv6. Grey cells beneath the robot constitute the reward map. Orange dashed line is the global path.}
    \label{fig:IRL}
\end{figure}

In this paper, we propose a novel socially-aware dynamic local planner SoLo T-DIRL using Trajectory-ranked Maximum-Entropy Deep Inverse Reinforcement Learning (T-MEDIRL)~\cite{T-MEDIRL}. Inspired by~\cite{SORO, SORL}, we extract features from the social dynamics between robots and pedestrians and unknown obstacle information in front of the robot. After collecting demonstrations by an operator, we use the sudden velocity change as a novel ranking score to address the sub-optimality in human demonstrations.

The proposed planning pipeline can take social dynamics and complex environmental information into account.

In particular, this work has the following contributions.
\begin{enumerate}
    \item A novel socially-aware T-MEDIRL-based local planner with multiple feature layers that can handle complex, indoor dynamic and crowded scenarios.
    \item A novel trajectory ranking score using the sudden velocity change of pedestrians around the robot to address the sub-optimality in human demonstrations. 
    \item The system has been evaluated and implemented in a ROS \cite{Quigley2009ROSAO} pipeline and is available for download at \href{https://github.com/UMich-CURLY/Fetch_IRL/}{https://github.com/UMich-CURLY/Fetch\_IRL}
\end{enumerate}

\section{Related Work}
\label{sec:relatedwork}
\subsection{Socially-aware Navigation}
In social environments, it is not sufficient for robots to plan a collision-free path~\cite{park2011smooth,HuVa}. The navigation behaviors should consider the context of interacting with humans, such as social norms~\cite{johnson2018socially}. In support of this claim, \cite{social_force_planner} proposes a model-based method that employed the social force model~\cite{social_force} in robot navigation. In~\cite{proactive_model}, besides considering the pedestrian model, the authors also takes interactive social information about human–objects and human group interactions into planner design. Recent methods of local planner design such as~\cite{CADRL, SAPP, SANN} use a socially-aware reward function in a deep reinforcement learning framework to accomplish planning tasks while maintaining a social distance away from humans. However, one of the limitations of these methods is that they fail to integrate other unknown obstacles (e.g., chairs and tables) in the environment into their reward function design. Another downside is the use of a single-feature reward function, making their performance highly dependent on the accuracy of human detection in real-life applications. To overcome these limitations, we propose a multi-modal local planner that considers both the socially-aware factors and static as well as dynamic obstacle information.
\subsection{IRL-based Navigation}

Instead of using a handcrafted reward function, MEDIRL-based methods generate rewards from demonstrations that can overcome the inaccuracy and incompleteness of forming rewards by hand. This is especially useful for social navigation problem where handcrafting a reward for interaction is hard.
\cite{MEIRL} presents MEIRL based on the principle of maximum entropy~\cite{ME} and IRL~\cite{IRL} to find the policy with the highest entropy subject to feature matching. \cite{SORL} proposes the structure of the IRL-based local planner in a crowded environment. \cite{SORO} successfully combines a learned human cooperative navigation model and MEDIRL into socially compliant mobile robot navigation. However, the IRL-based methods highly rely on the quality of human demonstrations, resulting in bad navigation performance if the latter are not optimal. \cite{T-MEDIRL} proposes an energy-based trajectory ranking loss in the training process of MEDIRL to eliminate the effect of non-optimal and sub-optimal demonstrations. For our proposed method, we extend the usage of T-MEDIRL to the socially-aware navigation field using the sudden velocity change of nearby people as the trajectory ranking loss. As a result, our navigation model not only considers socially-aware factors but is also less reliant on the quality of demonstrations.

\subsection{Human Trajectory Prediction}

In this work, a human trajectory prediction module is used as one of our feature layers to navigate the crowd safely and plan a socially acceptable path. Some physics-based methods generate the predicted trajectory by modeling people's motion using Newton's second law~\cite{Survey}. \cite{linear_model} predicts using a linear walking model and current velocity. \cite{Cyclist} uses a cyclist dynamic model which contains driving force and resistant force from acceleration, inclination, rolling, and air to predict people's trajectory. However, these model-based methods rely on the model's accuracy and ignore the randomness of people's walking speed and directions. Recently, a pattern-based method that can predict the trajectory by discovering statistical behavioral patterns from the pedestrian walking dataset is becoming popular. \cite{social_gan} proposed to predict trajectory by training adversarially against a recurrent discriminator. We use Social LSTM~\cite{Social_LSTM} in one of our feature layers to learn general human movement and predict their future trajectories for our proposed navigation method.

This work creates a dynamic multi-modal local planner that can handle socially-aware navigation tasks in a crowded environment. The feature map used in our local planner encodes different social dynamic information containing social distance and people's walking intentions. Moreover, we choose sudden velocity change as the trajectory ranking loss added to the training process to rank the demonstrations and improve the performance of our reward model.
\begin{figure*}[t]
    \centering
    \includegraphics[width = 0.75\textwidth]{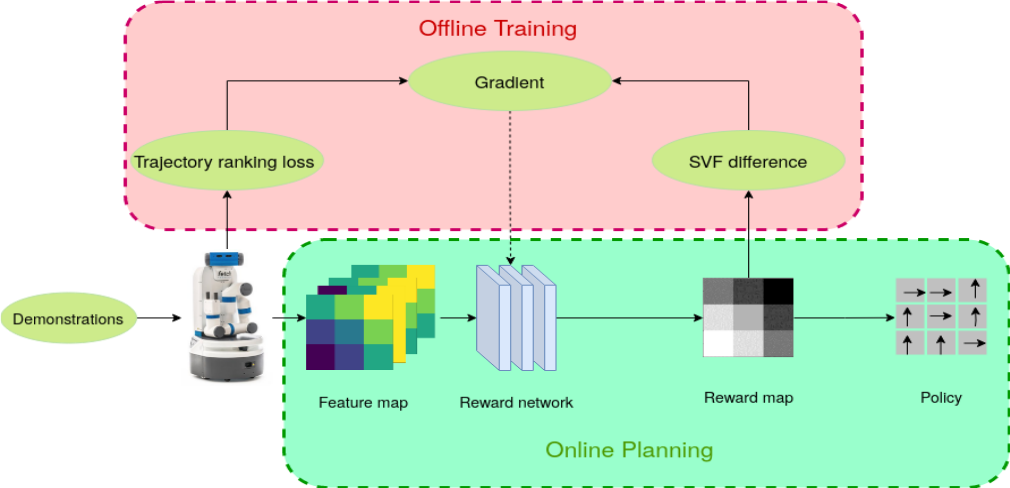}
    \caption{The structure plot of the system. From the plot, our system is divided into training and planning parts. The gradient for training is constructed using trajectory ranking loss and SVF difference. The reward network we use is a 3-layer fully-connected network whose input is feature map and the output is reward map.}
    \label{fig:system_structure}
\end{figure*}

\section{Methodology}
\label{sec:methodology}

We separate our work into offline training and online planning. We construct a multi-layer feature map from sensory information in the training part. Then we set different navigation goals and navigate the robot through the crowd to get demonstrations. After getting enough demonstrations, we apply the T-MEDIRL algorithm to train our IRL model until convergence. In the planning process, we get the reward for each state for policy and value iteration to get a local policy inside the grid. We then execute the policy and navigate the crowd using a PID controller. 

\subsection{Problem Statement}
\label{subsec: problem_statement}

We model the navigation problem as a three-layer process. The first layer is a planner generating the order of navigation goals that take the shortest time and distance by solving an optimization problem~\cite{Fu_2021}. The second layer is a global planner outputting the waypoints from the robot to each goal using a Voronoi-based planner \cite{PCR}. The third layer is our dynamic, socially aware local planner, which controls the robot going through each waypoint while avoiding any collisions with all obstacles and humans. 

The navigation structure of our local planner can be modeled as a Markov Decision Processes (MDP). An MDP can be defined as a tuple $\mathcal{M} = \{\mathcal{S,A,T},r,\gamma\}$, where $\mathcal{S}$ represents the state space of the system, $\mathcal{A}$ represents the action space, $\mathcal{T}$ is the transition probability, $r$ is the reward function and $\gamma\in[0,1)$ is the discount factor. The objective of the MDP is to find an optimal policy $\pi^{*}:\mathcal{S\rightarrow A}$ that maximizes the expected future reward:
\begin{equation}
    \pi^{*} = \argmax~ \mathbb{E}\left[\sum_{t=0}^{\infty}\gamma^{t}r\left(t \mid \pi \right)\right] .
    \label{eq: cost function}
\end{equation}

For our problem setting, the state space, $\mathcal{S}$, is defined as local grid cells in front of the robot. Each cell in the grid cells is one state in the state space. Thus, the number of states equals the number of local grid cells. For example, for a $m \times m$ grid, we will have $m^{2}$ states. The action space, $\mathcal{A}$, is defined as a set of discrete actions that move from one cell to its adjacent cells, i.e., $\mathcal{A}:=\{\uparrow, \downarrow, \leftarrow, \rightarrow, stop \}$. The action set can be easily extended to eight actions, including diagonal directions. The transition probability $\mathcal{T}(s,a,s')$ is deterministic in our problem setting. In \eqref{eq: reward function}, the reward of agents passing from one state to another is calculated using a fully-connected neural network $f$ with a set of parameter $\theta$ whose input is the feature vector $\phi$ from that state. We define the feature vectors $\phi: \mathcal{S\times A}\rightarrow\mathbb{R}^{n}$ that map a state and an action to an n-dimensional feature vector. 
\begin{equation}
    r(s,a)=f(\phi(s,a);\theta) .
    \label{eq: reward function}
\end{equation}


\subsection{T-MEDIRL}
\label{subsec: T-MEDIRL}

Maximum Entropy Deep Inverse Reinforcement Learning is used to find a set of weights that maximize the total reward of demonstrations. The IRL problem can be framed as MAP estimation, which maximizes the joint posterior distribution of observing expert demonstrations, $\mathcal{D}$, under a given reward structure and of the model parameters $\theta$ \cite{MEDIRL}. The joint log-likelihood $\mathcal{L}$ is given by \eqref{eq: likelyhood}.
\begin{equation}\label{eq: likelyhood}
    \begin{aligned}
    \mathcal{L}(\theta)=\log P(\mathcal{D}, \theta|r)&=\log P(\mathcal{D}|r)+\log P(\theta)  \\
    &=\mathcal{L}_{\mathcal{D}} + \mathcal{L}_{\theta} .
    \end{aligned}
\end{equation}

Equation~\eqref{eq: likelyhood} contains a data term $\mathcal{L}_{\mathcal{D}}$ and a model regulariser $\mathcal{L}_{\theta}$. After applying the chain rule, the MEDIRL gradient can be written as \eqref{eq: gradient}

\begin{equation}
    \label{eq: gradient}
    \begin{aligned}
    \frac{\partial\mathcal{L}}{\partial\theta} &= \frac{\partial\mathcal{L}_{\mathcal{D}}}{\partial\theta} + \frac{\partial\mathcal{L}_{\theta}}{\partial\theta} 
    =\frac{\partial\mathcal{L}_{\mathcal{D}}}{\partial r} \cdot \frac{\partial r }{\partial\theta} + \frac{\partial\mathcal{L}_{\theta}}{\partial\theta} \\
    &= (\mu_{\mathcal{D}}-\mathbb{E}[\mu]) \cdot \frac{\partial r }{\partial\theta} + \frac{\partial\mathcal{L}_{\theta}}{\partial\theta} .
    \end{aligned}
\end{equation}
$\mu_{\mathcal{D}}$ is the average State Visitation Frequencies (SVF) calculated from the training data. $\mathbb{E}[\mu]$ is the expected SVF calculated from the current prediction model. 

Note that in the existing MEDIRL framework, the gradient for training can only come from the difference between the SVF of demonstrations and the current prediction model. Hence, it highly relies on the quality of demonstrations. Inspired by \cite{T-MEDIRL}, we add a trajectory loss term to the MEDIRL framework to solve this problem. As shown in Fig.~\ref{fig:system_structure}, given a set of demonstrations, the objective is to find a policy with higher rewards for the high-ranked demonstrations. Instead of using one demonstration to train in one epoch, we use two demonstrations randomly chosen from the training dataset and add a pair-wise trajectory ranking loss as 

\begin{equation}
    \label{eq: trajectory loss rank}
    \begin{aligned}
    \mathcal{L}_{ij}=-\sum_{r_{i} < r_{j}}\log\frac{\exp r_{j}}{\exp r_{i} + \exp r_{j}} ,
    \end{aligned}
\end{equation}
where, $r_{i}, r_{j}$ are the trajectory rewards for trajectories $i, j$ respectively. The trajectory reward is reliant on the behaviour of the robot when it passes through the crowd since lack of social-awareness can cause disruptions among the crowd which are hard to define and measure quantitatively. To extrapolate beyond the sub-optimal demonstrations and take the negative effects into our trajectory loss design, we propose to use the Sudden Velocity Changes Rate (SVCR) of each demonstration for trajectory ranking in the T-MEDIRL framework. The SVCR, $\epsilon_{s}$, is defined as the number of persons, $n_{s}$, inside the grid cells whose velocity changes exceed a threshold divided by the length of trajectory $l_{R}$, i.e., $\epsilon := \frac{n_{s}}{l_{R}}$. 


For a crowded environment, using SVCR can help us rank robot trajectories that are disruptive to the humans in the scene and avoid reinforcing that behavior to our IRL agent. 
The algorithm for calculating SVCR is shown in Algorithm~\ref{alg:SVCR}, and Table~\ref{tab:variable definition} shows the variables definitions.

\begin{table}[t]
    \centering
        \caption{Variable Definition.}
    \footnotesize
    \resizebox{\columnwidth}{!}{\begin{tabular}{llll}\toprule
     \bf Variable & \bf Definition & \bf Variable & \bf Definition \\ \midrule
        \textbf{v}$_{t,n}$ & \begin{tabular}{@{}l@{}}n-th pedestrian's linear velocity \\ at time step t, \textbf{v}$_{t,n}\in\mathbb{R}^{2}$\end{tabular}  & $\omega_{t,n}$ &         \begin{tabular}{@{}l@{}}n-th pedestrian's angular velocity \\ at time step t, $\omega_{t,n}\in\mathbb{R}^{2}$\end{tabular}             \\ \midrule
        
        $v_{\text{thrd}}$ & \begin{tabular}{@{}l@{}}Linear velocity threshold for  \\ judging velocity sudden change\end{tabular} &   $\omega_{\text{thrd}}$   &   \begin{tabular}{@{}l@{}}Angular velocity threshold for  \\ judging velocity sudden change\end{tabular}\\ \midrule
        $S$ &       \begin{tabular}{@{}l@{}}The metric space of grid cells \\ in current demonstration \end{tabular} &            $l_{R}$               &  \begin{tabular}{@{}l@{}}The length of trajectory\\ in current demonstration\end{tabular}   \\ \midrule
        $N$ & Total number of pedestrians &         $\epsilon_{s}$          &  The velocity sudden change rate  \\ \midrule
        $n_{s}$ &  \begin{tabular}{@{}l@{}}The number of persons having sudden \\ velocity change within grid cells \end{tabular} & $\beta$ &\begin{tabular}{@{}l@{}}The gradient factor to control\\ the steepness of the function\end{tabular}  \\ \midrule
        $\phi(i,j)$ & The feature value of cell at (i,j) & $\alpha$ & Adjustment factor\\ \midrule
        $\rho_{\text{den}}$ & \begin{tabular}{@{}l@{}}The crowd density around\\ one person within 0.2m  \end{tabular}  & $d_{ij}$ &  \begin{tabular}{@{}l@{}}The distance between the cell\\ and the nearest person \end{tabular} \\ \midrule
        $\gamma$ & \begin{tabular}{@{}l@{}}Discount factor for predicted\\ trajectory feature \end{tabular}   & $d_{\text{social}}$ &  \begin{tabular}{@{}l@{}}The social distance of\\ the nearest person  \end{tabular} \\ \midrule
        $t$ & \begin{tabular}{@{}l@{}}The order of cells that\\  the trajectory extended to\end{tabular} &   $d_{ij}$ &  \begin{tabular}{@{}l@{}}The distance between the \\ cell and the nearest person\end{tabular}\\ \midrule
        $\textbf{x}_{t,n}$ & n-th pedestrian's position at time step $t$, $x_{t,n}\in\mathbb{R}^{2}$ \\ \bottomrule

    \end{tabular}}
    \label{tab:variable definition}
\end{table}

\begin{algorithm}[t]
\caption{Sudden Velocity Change Rate}\label{alg:SVCR}
\footnotesize
\textbf{Input:} $S,l_{R},\textbf{x}_{t,n}, v_{\text{thrd}}, \omega_{\text{thrd}}, \textbf{v}_{t,n}, \omega_{t,n},\textbf{v}_{(t-1),n}, \omega_{(t-1),n}$ \\
\textbf{Output:} $\epsilon_{s}$ 
\begin{algorithmic}
\State $n_{s}\leftarrow0$ \Comment{Initialization}
\For{ $t$ = 2 to $T$}
\For{ $n$ = 1 to $N$} 
\State $\Delta \textbf{v}_{t,n} \leftarrow \textbf{v}_{(t-1), n} - \textbf{v}_{t, n}$
\State $\Delta \omega_{t,n} \leftarrow \omega_{(t-1), n} - \omega_{t, n}$ \Comment{Calculate linear and angular velocity change}
\If{$\textbf{x}_{t,n}$ in $S$} \Comment{Whether inside grid cells}
\If{$||\Delta\textbf{v}_{t,n}|| \geq v_{\text{thrd}}$ \textbf{or}$ ||\Delta\omega_{t,n}|| \geq \omega_{\text{thrd}}$} \Comment{Whether velocity changes exceed threshold}
\State $n_{s}\leftarrow n_{s} + 1$
\EndIf
\EndIf
\EndFor 
\EndFor \\
\Return $\epsilon_{s} \leftarrow \frac{n_{s}}{l_{R}}$ \Comment{Normalization}

\end{algorithmic}
\end{algorithm}

\subsection{Features Extraction}
\label{subsec: feature_extraction}
LiDAR and RGB-D camera can detect obstacle information in the dynamic environment. We use this information to construct four distinct feature layers: distance to the goal, location of unknown static obstacles, predicted trajectory of people, and social distance feature layer. Besides the first two features, which can help the robot navigate to the goal without colliding with unknown obstacles, we use the other two social-awareness features to make the robot take into account social dynamics. 

\subsubsection{Distance to Goal feature}
\label{subsubsec: distance to goal}

The distance to goal feature is calculated between each cell and the nearest waypoint. After normalization, this feature can be used to propose the direction of navigation for the robot. 

\subsubsection{Unknown Obstacle feature}
\label{subsubsec: static obstacle information}

From the LiDAR information, we can extract the unknown obstacle feature by using Breshenham’s Algorithm. This feature can help the robot avoid the static obstacles other than people. 

\subsubsection{Predicted Trajectory feature}
\label{subsubsec: predicted trajectory}
Since humans are unlikely to stand still in a public environment, adding their walking intention and making the robot aware of the walking direction are crucial to social navigation. The current IRL-based method only uses the velocity and walking direction of the nearby people in the feature map as a criterion for judging their walking intention~\cite{SORO}. However, this feature is based on the constraint that people walk at a constant speed and fixed direction. To relax the constraints and fully use the past trajectories and social context, we propose using Social-LSTM~\cite{Social_LSTM} to predict these trajectories and encode the trajectory information into one of our feature layers. 

\begin{figure}
    \centering
    \includegraphics[width =0.48\textwidth]{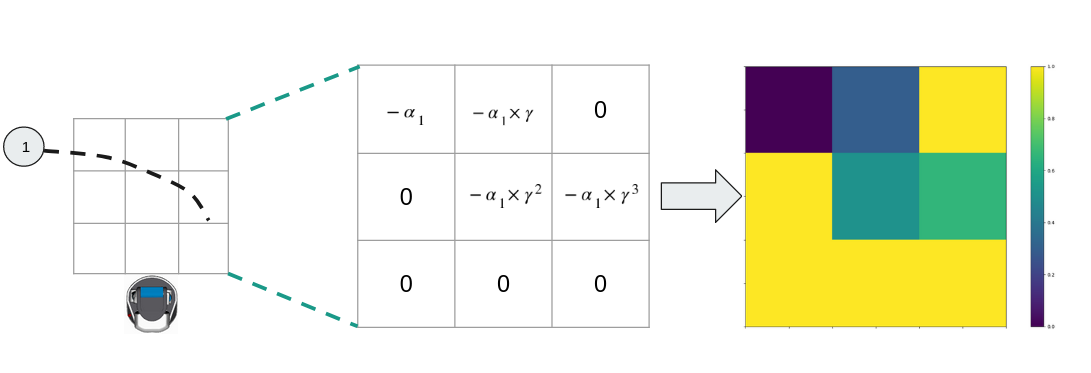}
    \caption{The trajectory prediction feature layer. The grey circle on the left figure is one pedestrian about to pass through the grid cells in front of the robot. The dash line is the trajectory predicted from social LSTM. The middle figure contains the raw feature values calculated according to the predicted trajectory. The figure on the right is the heatmap of the normalized feature layer.
    }
    \label{fig:prediction_feature}
\end{figure}

The spatio-temporal feature layer value is calculated according to the predicted trajectories from Social-LSTM algorithm. As shown in Fig. \ref{fig:prediction_feature}, instead of setting the binary value to each cell, we add a discount factor according to the order of cells to which the trajectories extended. We calculate the prediction feature of each cell using 
\begin{equation}
    \label{eq: trajectory feature}
    \begin{aligned}
    \phi_{ij} = -\sum_{n=1}^{N}\alpha_{nij}\cdot\gamma^{t} ,
    \end{aligned}
\end{equation}
\begin{equation}
    \label{eq: alpha value}
    \begin{aligned}
    \alpha_{nij} =
    \begin{cases}
      1 & \text{\small if the trajectory of n-th pedestrian in cell$(i,j)$}\\
      0 & \text{\small otherwise}
    \end{cases}       .
    \end{aligned}
\end{equation}
See TABLE~\ref{tab:variable definition} for the variable definitions.

\subsubsection{Social Distance feature}
\label{subsubsec: social distance feature}
Social distance is the minimum physical distance between two people who feel comfortable in a social context \cite{Social_distance}. Keeping a good social distance between robots and people is vital for socially aware navigation. However, most RL-based methods \cite{SAMP} consider the social distance as fixed, which can cause freezing-robot problem \cite{trautman_krause_2010}. Inspired by \cite{SAPP}, we encode a resilient social distance feature layer into our feature map for training. Since people are more likely to be comfortable with others coming closer in a high-density environment than in a low-density environment, instead of treating social distance as a fixed value, we treat the social distance of a person as a function of the surrounding crowd density as
\begin{equation}
    \label{eq: distance feature}
    d_{\text{social}} = \frac{1.577}{(\rho_{\text{den}}-0.8824)^{0.215}} - 0.967 , 
\end{equation}

\begin{figure}
    \centering
    \includegraphics[width =0.48\textwidth]{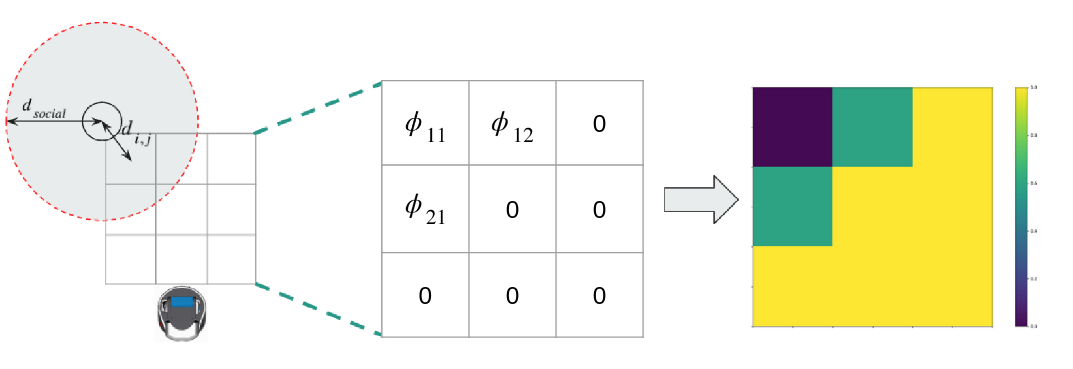}
    \caption{ The social distance feature layer. The big grey circle with red dash line boundary is the social comfortable area of the person. The middle grid cells represent the calculated feature layer. The figure on the right is the heatmap of the normalized feature layer. $\phi_{ij}=\alpha\times\frac{d_{i,j}^{\beta}-d_{\text{social}}^{\beta}}{d_{\text{social}}^{\beta}}$.
    }
    \label{fig:social_distance}
\end{figure}

\cite{SAPP} derives this equation by doing a curve fitting over the Asia and Pacific Trade Center (ATC) dataset \cite{ATCdataset}. As shown in Fig.  \ref{fig:social_distance}, we add a social distance feature according to the distance between individuals in the crowd and the robot as well as the social distance of the people. In order to take the social distance into our reward map, the feature value function can be separated into two cases: one is inside the social area, and the other is outside it. The feature value of each grid cell is calculated below:
\begin{equation}
    \label{eq: trajectory loss}
    \begin{aligned}
    \phi_{ij} =
    \begin{cases}
      \alpha\times\frac{d_{ij}^{\beta}-d_{\text{social}}^{\beta}}{d_{\text{social}}^{\beta}} & \text{$d_{t} \leq d_{\text{social}}$}\\
      0 & \text{$d_{t} > d_{\text{social}}$}
    \end{cases}  
    \end{aligned}
\end{equation}

\section{Results and Discussion}
\label{sec:results}

Our evaluation can be divided into two parts: socially aware feature evaluation and trajectory ranking loss evaluation. Sec.~\ref{subsec: simulation setup} introduces our simulation environment and training dataset. Sec.~\ref{subsec: social eval} evaluates our socially aware feature map by comparing our method with other socially aware navigation methods. In Sec.~\ref{subsec: traj loss eval}, we evaluate the trajectory ranking loss by comparing our method (SoLo T-DIRL) with the original MEDIRL-based method (SoLo DIRL).

\subsection{Training and Simulation Setup}
\label{subsec: simulation setup}

For the training of SoLo T-DIRL, we collect our data in a Gazebo environment with several unknown obstacles and random people walking around. The moving pedestrian simulator is the open source PedSim simulator~\cite{PedSim} which uses the social-force model to imitate pedestrians. The pedestrians in that simulator have virtual sensors and can avoid dynamic obstacles using the social-force model. 

We randomly choose several different goals and generate more than 100 demonstrations. We can generate feature maps and their corresponding trajectories according to the demonstration data and different sensor information from LiDAR, and an RGB-D camera. Our feature map size for training is $3\m \times 3\m$. When the robot moves through different grid cells, we collect the number of sudden velocity change of each demonstration for calculating trajectory ranking loss in~\eqref{eq: trajectory loss rank}. The trajectory of each demonstration is also collected for calculating SVCR combined with the number of sudden velocity change. The size of grid cells is $3 \times 3$ whose resolution is $1 \m/\mathrm{cell}$, and there is no overlapping between two different grid cells.

\subsection{Socially-aware Features Evaluation}
\label{subsec: social eval}
To evaluate the socially aware factors, we let the robot and people do a circle crossing in an open area, where
all the humans and robot are randomly positioned on a circle of radius $4\m$ and their goal positions are on the opposite side of the same circle. In this scenario, the robot learns to avoid taking risks to pass through the crowd and learns to keep a comfortable social distance.

\begin{figure}
    \centering
    \includegraphics[width =0.48\textwidth]{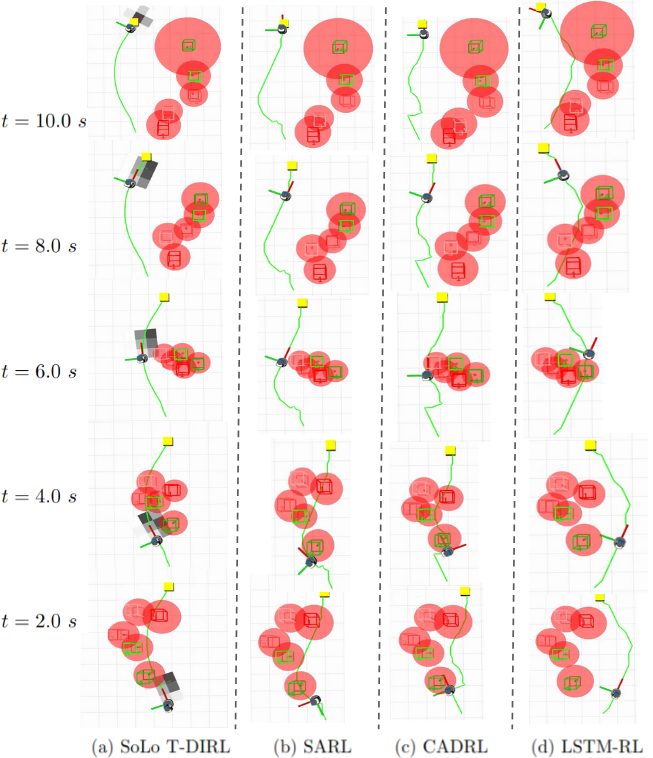}
    \caption{The comparison of SoLo T-DIRL with other baseline models in a circle crossing simulation environment. From the first row to last row: SoLo T-DIRL (Ours), SARL\cite{SARL}, CADRL\cite{CADRL}, LSTM-RL\cite{LSTM_RL}. The yellow block is the goal position and the green line is the trajectory of the robot. Each column represents the navigation process using the corresponding method. The navigation process is from bottom to up.} 
    \label{fig: qualitative_analysis}
\end{figure}

We compare our model with several other socially aware navigation methods widely used in the literature: SARL, CADRL, and LSTM-RL. As shown in Fig.~\ref{fig: qualitative_analysis}, the navigation path of different methods is compared where the pedestrian paths are similar for an exact comparison. Among these methods, our navigation trajectory generated by our SoLo T-DIRL method is the smoothest and the safest. Because we use the trajectory prediction as one of our feature layers, the robot will choose to pass the crowd on the opposite side of the walking direction of people. For example, if people are walking to the right side of the scene, our robot will choose to walk through from the left side. Also, since our social distance is also taken into our feature map, our robot will tend to keep a proper social distance from people. However, when encountering people in the center space, CADRL passes through them aggressively, which can cause invasion of social areas of pedestrians and even collision. LSTM-RL behaves in a risky manner by walking through from the right side of the scene, which may cause a collision by interfering in people's way. When encountering people in the front, SARL changes its orientation suddenly to avoid the crowd, which causes a delay in navigation time. 

\begin{table}[t]
    \centering
        \caption{Quantitative results of socially aware features. ``Time'' means the average time spent on the navigation process. ``Success'' means the rate at which the robot reaches the goal without collision. ``Invasion'' means the number of invasions into the social distance per meter.}
    \footnotesize
    \begin{tabular}{cccc}\toprule
     Method & Time & Success & Invasion \\ \midrule
     SoLo T-DIRL (Ours) & \textbf{10.41s} & \textbf{100\%} & \textbf{0.0112} \\ \midrule
     SARL~\cite{SARL} & 10.58s & 100\% & 0.0235 \\ \midrule
     CADRL~\cite{CADRL} & 10.82s & 94\% & 0.1202 \\ \midrule
     LSTM-RL~\cite{LSTM_RL} & 11.29s & 98\% & 0.0627 \\ \bottomrule
    \end{tabular}
    \label{tab:quantity eval}
\end{table}

In TABLE~\ref{tab:quantity eval}, we compare our method with SARL, CADRL, and LSTM-RL in four ways. Invasion rate is defined as the number of invasions into people's social area circle, whose radius is the social distance, divided by the total trajectory length. From the table, we know our method exceeds the other three methods as expected. Because our robot can predict people's intentions and plan a safe path, it will take fewer risks, saving time and increasing the success rate. Also, our method outperforms the baselines for invasion rate by adding a flexible social distance feature to our feature map. However, SARL and CADRL use the fixed social distance (0.2 $\m$) in their reward function, which can cause the robot to invade people's social area when the social distance is more than 0.2 $\m$. LSTM-RL will slow down when the robot approaches people, increasing the navigation time.

\subsection{Trajectory Ranking Loss Evaluation}
\label{subsec: traj loss eval}

Besides the feature map, we also evaluate the effect of our trajectory ranking loss in reducing the influence of suboptimal demonstrations. For comparison, we train our network without adding trajectory ranking loss as the baseline. In the classical IRL problem, the model's performance can be evaluated by comparing the reward with the ground truth. However, it is difficult to know the ground truth reward in a real-world problem. Inspired by \cite{T-MEDIRL}, we use two different metrics for evaluation.

The first is the sudden velocity change rate. To evaluate whether SoLo T-DIRL can improve robot behavior in the social environment, we calculate SVCR in the test dataset. The test dataset is collected in different scenarios with a different number of people walking around. Besides maximizing the reward, SoLo T-DIRL is expected to minimize SVCR, leading to less disturbance during navigation.

The second is the accuracy of classification. The basic idea of using trajectory ranking loss is to choose better demonstrations using a criterion other than the reward function. The demonstrations with a higher trajectory loss should output a lower discounted cumulative reward. Randomly picking up two different demonstrations, if the discounted cumulative reward of the demonstration with higher trajectory ranking loss is less than the other, we say this pair is considered ``correct''. The accuracy for classification is defined as the number of correct pairs over the number of total pairs.

\begin{table}[t]
    \centering
        \caption{Quantitative results for trajectory ranking loss. We compare SoLo T-DIRL with the original SoLo DIRL method and human teleoperation. ``-'' means there is no meaning for getting accuracy from human observation.}
    \footnotesize
    \begin{tabular}{ccc}\toprule
     Method & Accuracy & SVCR \\ \midrule
     SoLo T-DIRL & \textbf{60.6\%} & \textbf{0.2538} \\ \midrule
     SoLo DIRL& 38.8\% & 0.2968 \\ \midrule
     Human& - & 0.2672 \\ \midrule
    \end{tabular}
    \label{tab:loss eval}
\end{table}

From TABLE~\ref{tab:loss eval}, our SoLo T-DIRL both does a better job in accuracy and SVCR evaluation, which shows the effect of trajectory ranking loss on reward regulation. It is worth mentioning that SoLo T-DIRL does a better job decreasing SVCR than human teleoperation because people may do some suboptimal demonstrations in the navigation process and cause high SVCR. However, SoLo T-DIRL has put a lower weight on these suboptimal demonstrations, so the final training result is better than human teleoperation. 

\subsection{Limitations and Future Work}

This work relies on handcrafted features for learning our reward function. Although handcrafted features are explainable, it is impossible for us to consider all social dynamics. To model social dynamics, using deep neural networks to extract socially-aware features directly from sensor information \cite{DeepSocial2, DeepSocial} is a promising future direction. Furthermore, a hybrid approach where explainable features ensure the necessary inputs and raw sensory data enables lossless training is also an interesting future study.

The action space of our problem formulation is simplified and discrete, which does not consider the inertial dynamics of the robot. To incorporate the robot dynamics into the planning pipeline, the addition of inertial features to the feature layers~\cite{T-MEDIRL} or using Model Predictive Control (MPC)~\cite{teng2021toward,teng2022error} with robot dynamics constraints are interesting future research directions. In addition, the current grid map for planning is coarse and only covers a small area. In the future, we wish to improve the scalability of the proposed method by using finer resolutions and larger state space.

Finally, a real world experiment was carried out in a tight indoor environment as shown in Fig.\ref{fig:IRL}. We use YOLOv6~\cite{YOLOv6} for people detection and depth point cloud data for 3D projection. Due to hardware limitations and lack of filtering, the locations of people fluctuate dramatically. Moreover, with the lack of real world robot data as well as the gap between simulation and real life, the performance of the model is lower than the simulation. Therefore, to replicate the success of the simulation experiments on the real robot, more work on the perception algorithms is required. In particular, a reliable 3D dynamic object tracking can increase the success rate on hardware. Another possible direction is to develop a real-time world model that takes into account the dynamics and semantics of the scene~\cite{hughes2022hydra,9815141}.

\section{Conclusion}
\label{sec:conclusion}

We developed a socially-aware dynamic local planner based on the T-MEDIRL method to handle sub-optimal demonstrations. In particular, we proposed a novel trajectory ranking loss using the sudden velocity change rate. The ranking strategy provides a built-in value system for the robot to outperform expert demonstrations systematically. Our experiments showed promising results towards developing a socially assistive robot that can operate among people in everyday activities. We are also excited about exploring the role of language and visual cues in human-robot interactions within the context of social navigation as future work.

\clearpage

{\small 
\balance
\bibliographystyle{IEEEtran}
\bibliography{bib/strings-abrv,bib/ieee-abrv,bib/references}

\begin{thebibliography}{10}
\providecommand{\url}[1]{#1}
\csname url@rmstyle\endcsname
\providecommand{\newblock}{\relax}
\providecommand{\bibinfo}[2]{#2}
\providecommand\BIBentrySTDinterwordspacing{\spaceskip=0pt\relax}
\providecommand\BIBentryALTinterwordstretchfactor{4}
\providecommand\BIBentryALTinterwordspacing{\spaceskip=\fontdimen2\font plus
\BIBentryALTinterwordstretchfactor\fontdimen3\font minus
  \fontdimen4\font\relax}
\providecommand\BIBforeignlanguage[2]{{%
\expandafter\ifx\csname l@#1\endcsname\relax
\typeout{** WARNING: IEEEtran.bst: No hyphenation pattern has been}%
\typeout{** loaded for the language `#1'. Using the pattern for}%
\typeout{** the default language instead.}%
\else
\language=\csname l@#1\endcsname
\fi
#2}}

\bibitem{SARs}
D.~Feil-Seifer and M.~Mataric, ``Defining socially assistive robotics,'' in
  \emph{Proc. Int. Conf. Rehabilitation Robotics}, 2005, pp. 465--468.

\bibitem{museum_robot}
\BIBentryALTinterwordspacing
N.~Gasteiger, M.~Hellou, and H.~S. Ahn, ``Deploying social robots in museum
  settings: A quasi-systematic review exploring purpose and acceptability,''
  \emph{International Journal of Advanced Robotic Systems}, vol.~18, no.~6, p.
  17298814211066740, 2021. [Online]. Available:
  \url{https://doi.org/10.1177/17298814211066740}
\BIBentrySTDinterwordspacing

\bibitem{shopping_mall}
T.~Kanda, M.~Shiomi, Z.~Miyashita, H.~Ishiguro, and N.~Hagita, ``An affective
  guide robot in a shopping mall,'' in \emph{ACM/IEEE International Conference
  on Human-Robot Interaction}, 2009, pp. 173--180.

\bibitem{airport_robot}
R.~Triebel, K.~O. Arras, R.~Alami, L.~Beyer, S.~Breuers, R.~Chatila,
  M.~Chetouani, D.~Cremers, V.~Evers, M.~Fiore, H.~Hung, O.~A.~I. Ram{\'i}rez,
  M.~Joosse, H.~Khambhaita, T.~P. Kucner, B.~Leibe, A.~J. Lilienthal,
  T.~Linder, M.~Lohse, M.~Magnusson, B.~Okal, L.~Palmieri, U.~Rafi, M.~van
  Rooij, and L.~Zhang, ``Spencer: A socially aware service robot for passenger
  guidance and help in busy airports,'' in \emph{FSR}, 2015.

\bibitem{Eval_SARN}
Y.~Gao and C.-M. Huang, ``Evaluation of socially-aware robot navigation,''
  \emph{Frontiers in Robotics and AI}, vol.~8, 01 2022.

\bibitem{social_aware_navigation}
P.~Teja~Singamaneni, A.~Favier, and R.~Alami, ``Human-aware navigation planner
  for diverse human-robot interaction contexts,'' in \emph{Proc. {IEEE}/{RSJ}
  Int. Conf. Intell. Robots and Syst.}, 2021, pp. 5817--5824.

\bibitem{SORO}
B.~Kim and J.~Pineau, ``Socially adaptive path planning in human environments
  using inverse reinforcement learning,'' \emph{International Journal of Social
  Robotics}, vol.~8, pp. 51--66, 01 2016.

\bibitem{Fox1997TheDW}
D.~Fox, W.~Burgard, and S.~Thrun, ``The dynamic window approach to collision
  avoidance,'' \emph{{IEEE} Robot. Autom. Mag.}, vol.~4, pp. 23--33, 1997.

\bibitem{social_force_planner}
G.~Ferrer, A.~Garrell, and A.~Sanfeliu, ``Social-aware robot navigation in
  urban environments,'' in \emph{European Conference on Mobile Robots}, 2013,
  pp. 331--336.

\bibitem{SAPP}
\BIBentryALTinterwordspacing
X.~Lu, H.~Woo, A.~Faragasso, A.~Yamashita, and H.~Asama, ``Socially aware robot
  navigation in crowds via deep reinforcement learning with resilient reward
  functions,'' \emph{Advanced Robotics}, vol.~36, no.~8, pp. 388--403, 2022.
  [Online]. Available: \url{https://doi.org/10.1080/01691864.2022.2043184}
\BIBentrySTDinterwordspacing

\bibitem{CADRL}
\BIBentryALTinterwordspacing
Y.~F. Chen, M.~Liu, M.~Everett, and J.~P. How, ``Decentralized
  non-communicating multiagent collision avoidance with deep reinforcement
  learning,'' \emph{CoRR}, vol. abs/1609.07845, 2016. [Online]. Available:
  \url{http://arxiv.org/abs/1609.07845}
\BIBentrySTDinterwordspacing

\bibitem{SORL}
\BIBentryALTinterwordspacing
H.~Kretzschmar, M.~Spies, C.~Sprunk, and W.~Burgard, ``Socially compliant
  mobile robot navigation via inverse reinforcement learning,'' \emph{Int. J.
  Robot. Res.}, vol.~35, no.~11, pp. 1289--1307, 2016. [Online]. Available:
  \url{https://doi.org/10.1177/0278364915619772}
\BIBentrySTDinterwordspacing

\bibitem{T-MEDIRL}
L.~Gan, J.~W. Grizzle, R.~M. Eustice, and M.~Ghaffari, ``Energy-based legged
  robots terrain traversability modeling via deep inverse reinforcement
  learning,'' \emph{IEEE Robotics and Automation Letters}, vol.~7, no.~4, pp.
  8807--8814, 2022.

\bibitem{Quigley2009ROSAO}
M.~Quigley, K.~Conley, B.~Gerkey, J.~Faust, T.~Foote, J.~Leibs, R.~Wheeler,
  A.~Y. Ng, \emph{et~al.}, ``{ROS}: an open-source robot operating system,'' in
  \emph{ICRA workshop on open source software}, vol.~3, no. 3.2.\hskip 1em plus
  0.5em minus 0.4em\relax Kobe, Japan, 2009, p.~5.

\bibitem{park2011smooth}
J.~J. Park and B.~Kuipers, ``A smooth control law for graceful motion of
  differential wheeled mobile robots in 2d environment,'' in \emph{Proc. {IEEE}
  Int. Conf. Robot. and Automation}.\hskip 1em plus 0.5em minus 0.4em\relax
  IEEE, 2011, pp. 4896--4902.

\bibitem{HuVa}
T.~Kruse, A.~Kirsch, E.~A. Sisbot, and R.~Alami, ``Exploiting human cooperation
  in human-centered robot navigation,'' in \emph{19th International Symposium
  in Robot and Human Interactive Communication}, 2010, pp. 192--197.

\bibitem{johnson2018socially}
C.~Johnson and B.~Kuipers, ``Socially-aware navigation using topological maps
  and social norm learning,'' in \emph{Proceedings of the AAAI/ACM Conference
  on AI, Ethics, and Society}, 2018, pp. 151--157.

\bibitem{social_force}
D.~Helbing and P.~Molnar, ``Social force model for pedestrian dynamics,''
  \emph{Physical Review E}, vol.~51, 05 1998.

\bibitem{proactive_model}
X.-T. Truong and T.~D. Ngo, ``Toward socially aware robot navigation in dynamic
  and crowded environments: A proactive social motion model,'' \emph{{IEEE}
  Trans. Autom. Sci. Eng.}, vol.~14, no.~4, pp. 1743--1760, 2017.

\bibitem{SANN}
\BIBentryALTinterwordspacing
S.~Liu, P.~Chang, Z.~Huang, N.~Chakraborty, W.~Liang, J.~Geng, and
  K.~Driggs-Campbell, ``Socially aware robot crowd navigation with interaction
  graphs and human trajectory prediction,'' 2022. [Online]. Available:
  \url{https://arxiv.org/abs/2203.01821}
\BIBentrySTDinterwordspacing

\bibitem{MEIRL}
B.~D. Ziebart, A.~L. Maas, J.~A. Bagnell, and A.~K. Dey, ``Maximum entropy
  inverse reinforcement learning,'' in \emph{AAAI}, 2008.

\bibitem{ME}
\BIBentryALTinterwordspacing
R.~D. Rosenkrantz, \emph{Where Do We Stand on Maximum Entropy? (1978)}.\hskip
  1em plus 0.5em minus 0.4em\relax Dordrecht: Springer Netherlands, 1989, pp.
  210--314. [Online]. Available:
  \url{https://doi.org/10.1007/978-94-009-6581-2_10}
\BIBentrySTDinterwordspacing

\bibitem{IRL}
\BIBentryALTinterwordspacing
P.~Abbeel and A.~Y. Ng, ``Apprenticeship learning via inverse reinforcement
  learning.''\hskip 1em plus 0.5em minus 0.4em\relax New York, NY, USA:
  Association for Computing Machinery, 2004. [Online]. Available:
  \url{https://doi.org/10.1145/1015330.1015430}
\BIBentrySTDinterwordspacing

\bibitem{Survey}
A.~Rudenko, L.~Palmieri, M.~Herman, K.~Kitani, D.~Gavrila, and K.~Arras,
  ``Human motion trajectory prediction: a survey,'' \emph{Int. J. Robot. Res.},
  vol.~39, p. 027836492091744, 06 2020.

\bibitem{linear_model}
M.~Kollmitz, K.~Hsiao, J.~Gaa, and W.~Burgard, ``Time dependent planning on a
  layered social cost map for human-aware robot navigation,'' in \emph{2015
  European Conference on Mobile Robots (ECMR)}, 2015, pp. 1--6.

\bibitem{Cyclist}
S.~Zernetsch, S.~Kohnen, M.~Goldhammer, K.~Doll, and B.~Sick, ``Trajectory
  prediction of cyclists using a physical model and an artificial neural
  network,'' in \emph{IEEE Intelligent Vehicles Symposium (IV)}, 2016, pp.
  833--838.

\bibitem{social_gan}
A.~Gupta, J.~Johnson, L.~Fei-Fei, S.~Savarese, and A.~Alahi, ``Social gan:
  Socially acceptable trajectories with generative adversarial networks,'' in
  \emph{Proc. {IEEE} Conf. Comput. Vis. Pattern Recog.}, 2018, pp. 2255--2264.

\bibitem{Social_LSTM}
A.~Alahi, K.~Goel, V.~Ramanathan, A.~Robicquet, L.~Fei-Fei, and S.~Savarese,
  ``Social lstm: Human trajectory prediction in crowded spaces,'' \emph{Proc.
  {IEEE} Conf. Comput. Vis. Pattern Recog.}, pp. 961--971, 2016.

\bibitem{Fu_2021}
\BIBentryALTinterwordspacing
B.~Fu, T.~Kathuria, D.~Rizzo, M.~Castanier, X.~J. Yang, M.~Ghaffari, and
  K.~Barton, ``Simultaneous human-robot matching and routing for multi-robot
  tour guiding under time uncertainty,'' \emph{Journal of Autonomous Vehicles
  and Systems}, vol.~1, no.~4, oct 2021. [Online]. Available:
  \url{https://doi.org/10.1115%2F1.4053428}
\BIBentrySTDinterwordspacing

\bibitem{PCR}
\BIBentryALTinterwordspacing
T.~Kathuria, Y.~Xu, T.~Chakhachiro, X.~J. Yang, and M.~Ghaffari,
  ``Providers-clients-robots: Framework for spatial-semantic planning for
  shared understanding in human-robot interaction,'' 2022. [Online]. Available:
  \url{https://arxiv.org/abs/2206.10767}
\BIBentrySTDinterwordspacing

\bibitem{MEDIRL}
\BIBentryALTinterwordspacing
M.~Wulfmeier, P.~Ondruska, and I.~Posner, ``Maximum entropy deep inverse
  reinforcement learning,'' 2015. [Online]. Available:
  \url{https://arxiv.org/abs/1507.04888}
\BIBentrySTDinterwordspacing

\bibitem{Social_distance}
\BIBentryALTinterwordspacing
G.~A. Akerlof, ``Social distance and social decisions,'' \emph{Econometrica},
  vol.~65, no.~5, pp. 1005--1027, 1997. [Online]. Available:
  \url{http://www.jstor.org/stable/2171877}
\BIBentrySTDinterwordspacing

\bibitem{SAMP}
\BIBentryALTinterwordspacing
Y.~F. Chen, M.~Everett, M.~Liu, and J.~P. How, ``Socially aware motion planning
  with deep reinforcement learning,'' 2017. [Online]. Available:
  \url{https://arxiv.org/abs/1703.08862}
\BIBentrySTDinterwordspacing

\bibitem{trautman_krause_2010}
P.~Trautman and A.~Krause, ``Unfreezing the robot: Navigation in dense,
  interacting crowds,'' \emph{2010 IEEE/RSJ International Conference on
  Intelligent Robots and Systems}, 2010.

\bibitem{ATCdataset}
D.~Brščić, T.~Kanda, T.~Ikeda, and T.~Miyashita, ``Person tracking in large
  public spaces using 3-d range sensors,'' \emph{IEEE Transactions on
  Human-Machine Systems}, vol.~43, no.~6, pp. 522--534, 2013.

\bibitem{PedSim}
B.~Okal and K.~O. Arras, ``Towards group-level social activity recognition for
  mobile robots,'' 2014.

\bibitem{SARL}
\BIBentryALTinterwordspacing
C.~Chen, Y.~Liu, S.~Kreiss, and A.~Alahi, ``Crowd-robot interaction:
  Crowd-aware robot navigation with attention-based deep reinforcement
  learning,'' 2018. [Online]. Available: \url{https://arxiv.org/abs/1809.08835}
\BIBentrySTDinterwordspacing

\bibitem{LSTM_RL}
\BIBentryALTinterwordspacing
M.~Everett, Y.~F. Chen, and J.~P. How, ``Motion planning among dynamic,
  decision-making agents with deep reinforcement learning,'' 2018. [Online].
  Available: \url{https://arxiv.org/abs/1805.01956}
\BIBentrySTDinterwordspacing

\bibitem{DeepSocial2}
A.~Staroverov, D.~Yudin, I.~Belkin, V.~Adeshkin, Y.~Solomentsev, and A.~Panov,
  ``Real-time object navigation with deep neural networks and hierarchical
  reinforcement learning,'' \emph{IEEE Access}, vol.~8, pp. 195\,608--195\,621,
  01 2020.

\bibitem{DeepSocial}
X.~Yao, J.~Zhang, and J.~Oh, ``Following social groups: Socially-compliant
  autonomous navigation in dense crowds,'' November 2019.

\bibitem{teng2021toward}
S.~Teng, Y.~Gong, J.~W. Grizzle, and M.~Ghaffari, ``Toward safety-aware
  informative motion planning for legged robots,'' \emph{arXiv preprint
  arXiv:2103.14252}, 2021.

\bibitem{teng2022error}
S.~Teng, D.~Chen, W.~Clark, and M.~Ghaffari, ``An error-state model predictive
  control on connected matrix {Lie} groups for legged robot control,'' in
  \emph{Proc. {IEEE}/{RSJ} Int. Conf. Intell. Robots and Syst.}\hskip 1em plus
  0.5em minus 0.4em\relax IEEE, 2022, pp. 1--8.

\bibitem{YOLOv6}
\BIBentryALTinterwordspacing
C.~Li, L.~Li, H.~Jiang, K.~Weng, Y.~Geng, L.~Li, Z.~Ke, Q.~Li, M.~Cheng,
  W.~Nie, Y.~Li, B.~Zhang, Y.~Liang, L.~Zhou, X.~Xu, X.~Chu, X.~Wei, and
  X.~Wei, ``Yolov6: A single-stage object detection framework for industrial
  applications,'' 2022. [Online]. Available:
  \url{https://arxiv.org/abs/2209.02976}
\BIBentrySTDinterwordspacing

\bibitem{hughes2022hydra}
N.~Hughes, Y.~Chang, and L.~Carlone, ``Hydra: A real-time spatial perception
  system for {3D} scene graph construction and optimization,'' in \emph{Proc.
  Robot.: Sci. Syst. Conf.}, 2022.

\bibitem{9815141}
J.~Wilson, J.~Song, Y.~Fu, A.~Zhang, A.~Capodieci, P.~Jayakumar, K.~Barton, and
  M.~Ghaffari, ``{MotionSC}: Data set and network for real-time semantic
  mapping in dynamic environments,'' \emph{IEEE Robotics and Automation
  Letters}, vol.~7, no.~3, pp. 8439--8446, 2022.

\end{thebibliography}
}

\end{document}